\documentclass[10pt,conference]{IEEEtran} 
\IEEEoverridecommandlockouts
\usepackage{cite}
\usepackage{amsmath,amssymb,amsfonts}
\usepackage{algorithmic}
\usepackage{graphicx}
\usepackage{textcomp}
\usepackage[table]{xcolor}
\usepackage{multirow}
\usepackage{subfigure}
\usepackage{booktabs}
\usepackage{url}
\usepackage{marvosym}
\def\BibTeX{{\rm B\kern-.05em{\sc i\kern-.025em b}\kern-.08em
    T\kern-.1667em\lower.7ex\hbox{E}\kern-.125emX}}
\begin{document}

\bibliographystyle{IEEEtran}

% \title{Are Large Language Models Logical Solvers? \\ Solver Code Translation and Simulation with LLMs\\}

\title{Can Language Models Pretend Solvers?\\ Logic Code Simulation with LLMs}

\author{\IEEEauthorblockN{Minyu Chen\IEEEauthorrefmark{1}, Guoqiang Li\IEEEauthorrefmark{1}\,\Letter\thanks{\Letter\, Corresponding author},  Ling-I Wu\IEEEauthorrefmark{1}, Ruibang Liu\IEEEauthorrefmark{1}, Yuxin Su\IEEEauthorrefmark{1}, Xi Chang\IEEEauthorrefmark{2}, Jianxin Xue\IEEEauthorrefmark{2}}
\IEEEauthorblockA{\IEEEauthorrefmark{1}\textit{Shanghai Jiao Tong University, Shanghai 200240, China}\\
\{minkow, li.g, edithwuly, 628628, sshirley\}@sjtu.edu.cn
\\
\IEEEauthorrefmark{2}\textit{Shanghai Polytechnic University, Shanghai 201209, China}\\
\{changxi, jxxue\}@sspu.edu.cn}
}

% 虽然但是，逻辑真的会给人带来不幸...就没有叫logic solver的东西
% 因为logic大部分是用resolution，不是一种solver


%\author{\IEEEauthorblockN{Anonymous Authors}
% email address or ORCID
% }
% \and
% \and
% \and
% \and
% \and

\maketitle

\begin{abstract}

Transformer-based large language models (LLMs) have demonstrated significant potential in addressing logic problems. capitalizing on the great capabilities of LLMs for code-related activities, several frameworks leveraging logical solvers for logic reasoning have been proposed recently. While existing research predominantly focuses on viewing LLMs as natural language logic solvers or translators, their roles as logic code interpreters and executors have received limited attention. This study delves into a novel aspect, namely logic code simulation, which forces LLMs to emulate logical solvers in predicting the results of logical programs. To further investigate this novel task, we formulate our three research questions: Can LLMs efficiently simulate the outputs of logic codes? What strength arises along with logic code simulation? And what pitfalls? To address these inquiries, we curate three novel datasets tailored for the logic code simulation task and undertake thorough experiments to establish the baseline performance of LLMs in code simulation. Subsequently, we introduce a pioneering LLM-based code simulation technique, Dual Chains of Logic (DCoL). This technique advocates a dual-path thinking approach for LLMs, which has demonstrated state-of-the-art performance compared to other LLM prompt strategies, achieving a notable improvement in accuracy by 7.06\% with GPT-4-Turbo.

% Transformer-based large language models (LLMs) have demonstrated significant potential in addressing logic problems. capitalizing on the great capabilities of LLMs for code-related activities, several frameworks leveraging logical solvers for logic reasoning have been proposed recently. While existing research predominantly focuses on viewing LLMs as natural language logic solvers or translators, their roles as logic code interpreters and executors have received limited attention. This study delves into a novel aspect, namely logic code simulation, which encourages LLMs to emulate logical solvers in predicting the results of logical programs. Experiments results show LLMs' potential of the logic solver simulation task. Furthermore, we conduct a thorough evaluation of the interplay among logic code simulation, logic code comprehension, and logic code generation, offering fresh insights and implications for LLM-driven logic reasoning investigations. Additionally, we introduce a set of new prompts tailored for LLMs to enhance a variety of logic-solver-based applications.

\end{abstract}

% experts帮我看一下ref
\section{Introduction}
Logic serves as the foundation for the majority of formal methodologies in software engineering. By translating specifications, constraints, and test cases into logical frameworks like First Order Logic (FOL) or Satisfiability Modulo Theories (SMT), logic solvers can precisely judge arguments in problems or check the satisfiability of problems. Thus, state-of-the-art solvers including Z3~\cite{de2008z3} and cvc5~\cite{barbosa2022cvc5} are applied in wide aspects of software engineering, including software verification~\cite{cordeiro2011verifying}, software testing~\cite{soltana2020practical}, program synthesis~\cite{kang2019automated}, and program analysis~\cite{gadelha2019smt}. 
However, logic remains a complex subject within human cognition, presenting several challenges to the deployment of logic solvers in software engineering.  One key obstacle is the gap between natural language (NL) and the solver language (SL) of logical problems, impeding not only the programming process but also the comprehension of code segments within existing software systems. Besides, despite advancements, solvers encounter difficulties in effectively and precisely resolving logical problems, including those that are straightforward for humans. Furthermore, the integration of logic in software engineering often involves various extensions like arrays, integers, reals, strings, and bit-vectors. This diversity has led to the development of specialized solvers~\cite{gadelha2019smt, cai2022local} tailored to different logical scopes.

Recent advances in transformer-based large language models (LLMs) such as GPT~\cite{brown2020language} and LLaMA~\cite{touvron2023llama} have showcased their ability to perform logic reasoning akin to humans~\cite{pan2023logic, lee2024symba}. By incentivizing  LLMs to translate natural language into solver languages with in-context learning (ICL)~\cite{brown2020language} or model fine-tuning~\cite{feng2023language}, LLMs can successfully solve simple propositional logic questions in NL form~\cite{tafjord2021proofwriter, saparov2023language}. When faced with intricate logic-based queries in forms such as FOL~\cite{han2022folio} and SAT~\cite{zhong2022analytical}, LLMs exhibit challenges in direct problem-solving and show low successful execution rates of code generation. LLMs also fail to conduct relational reasoning~\cite {li2024llms}. 
Nonetheless, it is crucial to recognize the valuable bridge LLMs establish between natural language and solver language, along with the code intelligence LLMs bring to various aspects of software engineering, such as code generation~\cite{zhang2023infere, liu2023codegen4libs, yu2024codereval}, documentation~\cite{ma2024knowlog, xu2024unilog}, code understanding~\cite{wang2023generating}, and others~\cite{gupta2023grace, sun2023smt, deng2024large, yang2024large, sun2024gptscan}. 
On the other hand, a novel line of research, code simulation, is raised to discover the ability of LLMs to simulate the execution of code and algorithms. Preliminary research reports that the LLMs struggle to execute long and complex procedures~\cite{la2024code}.

\begin{figure}[!ht]
\centering
\includegraphics[width=8.5cm]{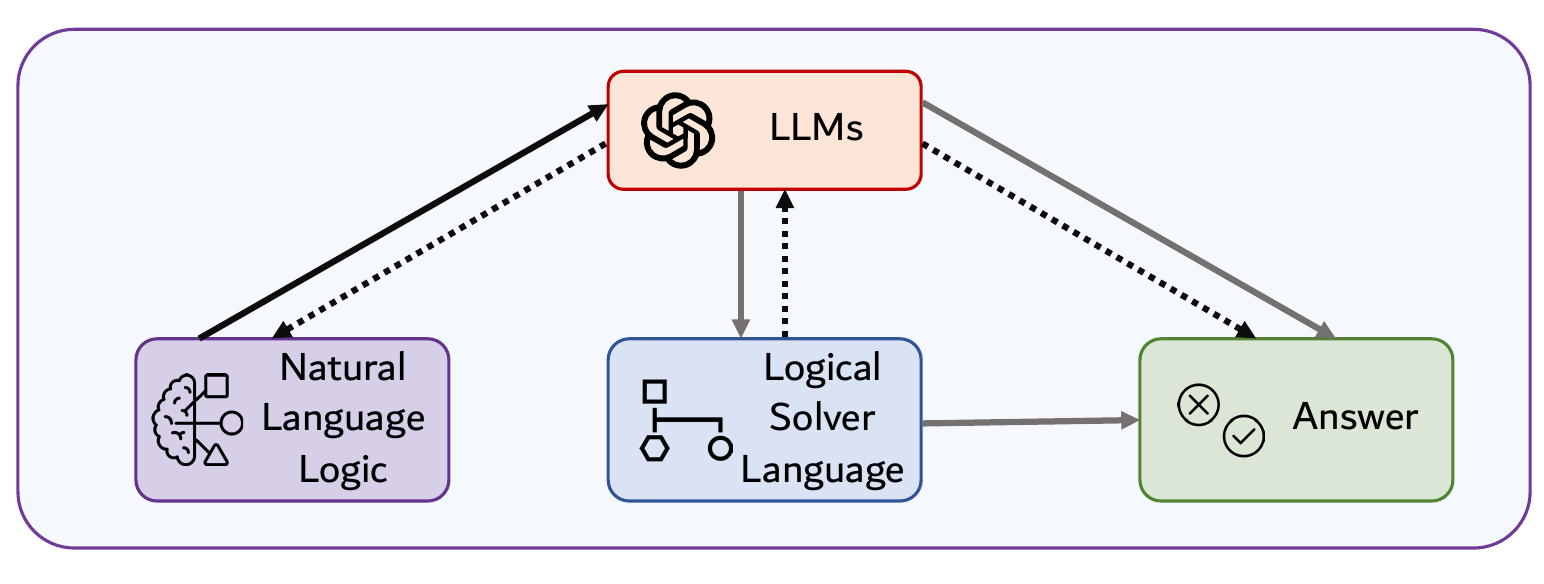}
\caption{An overview of concepts in our research. The solid line illustrates current research methods, encompassing two approaches to natural language problem-solving: LLM-based logic reasoning and solver-augmented LLM reasoning. Both methods leverage logic understanding with LLMs (indicated by the black solid line) but diverge in their reliance on logic solvers. The dotted lines represent crucial issues discussed in this paper that are not mentioned in previous studies.}
\label{fig1}
\end{figure}

\begin{figure*}[!t]
\centering
\includegraphics[width=16cm]{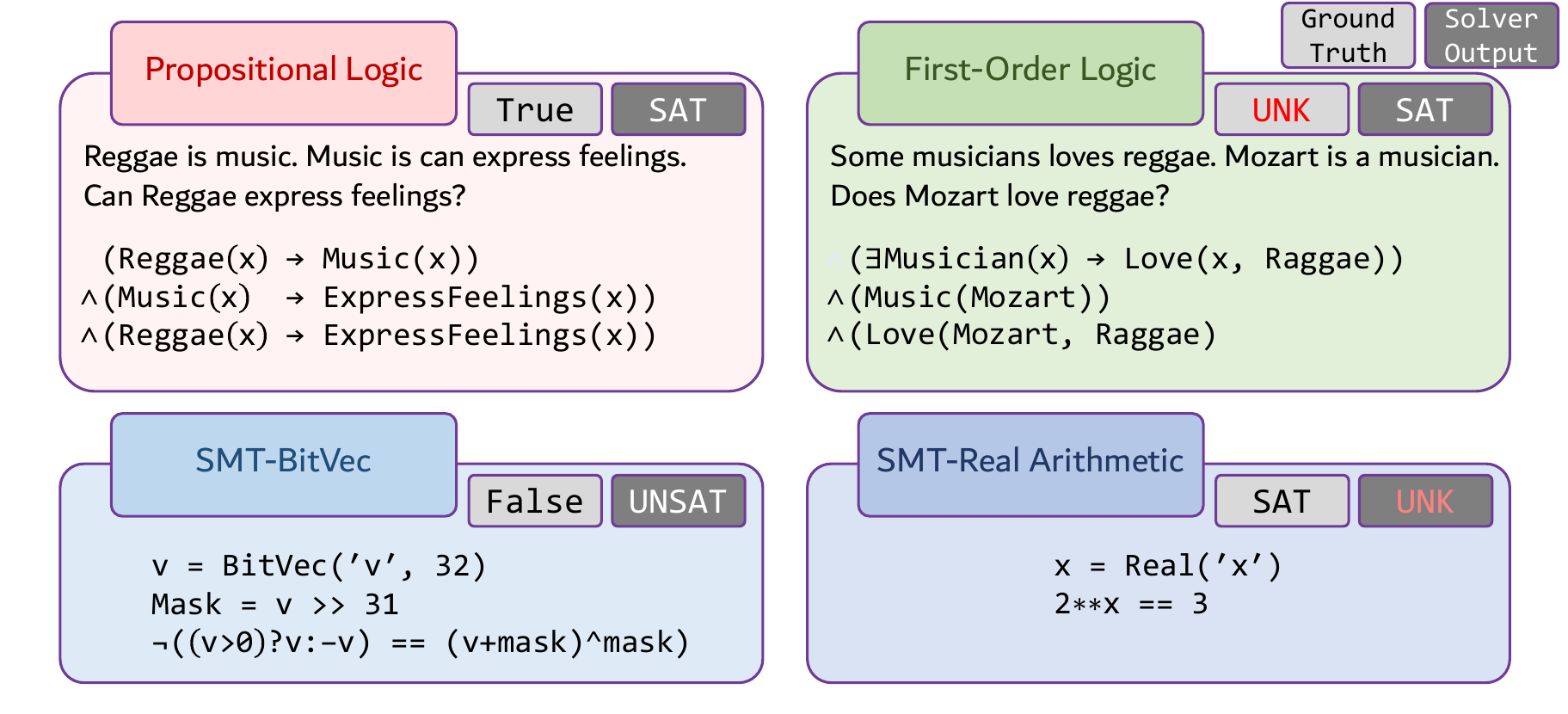}
\caption{Samples of logical problem to be studied in our research. The \textbf{light} gray boxes display the ground truth of the given problems, while the results given by the SMT solver are presented in the \textbf{dark} grey boxes. `UNK' denotes unknown.}
\label{fig2}
\end{figure*}

% 几个实验需要归类: 1. 直接模拟Z3代码(大，小数据集，Z3的缺陷） 2. mixed prediction(nl+sl，2个方向) 3. logic code 能不能被反向翻译回去 4. propmt调优 5. 多种solver语言uniform化成NL后做solve，并且支持如Z3 BAD CASE的 6 simulation+generation，先si再gen 7.代码运行的鲁棒性、多语言支持 8. chatgpt think on what？LANGUAGE?CODE execution?
This research addresses previously overlooked areas that have been left unexplored by prior studies. While previous works have predominantly focused on directly solving logic problems presented in natural language (NL) form or employing external logical solvers, Large Language Models (LLMs) have primarily served as NL reasoners or NL-SL translators in those endeavors.
Given that LLMs have been extensively utilized for logic reasoning in NL, it prompts the question: What about code simulation? As logic is always inherently implicit within code and is manifested through the code's execution, it serves as a fundamental component that guides the analysis process and influences the identification of defects, verification of correctness properties, and generation of test cases. To solve logic problems encoded within it, LLMs must comprehend implicit logic, engage in logic reasoning, and translate the results of reasoning back into the code's execution outcome, which constitutes the entire process of LLMs' simulation of logic codes. 

Thus, our study delves into the realm of LLMs' simulation of logic codes to ascertain whether LLMs can retain their proficiency in logic reasoning when applied to code as LLM-based logical solvers. 
To distinguish our work from others, we propose the first research question: \textit{\textbf{(RQ1)}Can LLMs efficiently simulate the outputs of logic codes?}
Through experiments, we demonstrate LLMs' competence in inferring the outcomes of codes that incorporate logical reasoning. Furthermore, we investigate the effectiveness of various prompt techniques, such as PS, CoSm, and COT, in enhancing LLMs' performance in simulating code.
To fully harness the potential of LLMs in logic code simulation, we introduce our framework, Dual Chains of Logic (DCoL), which stands as the first framework specifically tailored for LLM-based logic solvers. DCoL takes the code directly as input and guides LLMs to adopt a dual-path thinking approach, enabling them to draw conclusions for the code's result. This framework assists LLMs in effectively navigating through reasoning traps, such as multiple assignment problems, and reducing confusion during logic problem-solving.
Then, we raise the rest of our research questions: \textit{{\textbf{(RQ2)}}What strength and {\textbf{(RQ3)}}pitfalls arise along with logic code simulation?} 
To this end, we conduct a thorough examination of the effectiveness, strengths, and potential pitfalls of LLM-based logic solvers, offering a comprehensive analysis of their performance and identifying potential aspects for refinement and enhancement in the future.

% This research steps into overlooked areas left by previous studies. Prior works mostly focus on directly solving logic problems in natural language (NL) form or utilizing external logical solvers. LLMs perform as NL reasoners or NL-SL translators in those research. 
% Our research delves into LLMs' simulation of logic codes, aiming to investigate LLMs' capability of solving existing logic problems encoded in codes. 
% We conduct experiments to showcase LLMs' capability in inferring the outcomes of the codes with logic reasoning. Additionally, we investigated the efficacy of various prompt techniques, including PS, CoSm, and COT, in enhancing LLMs' performance in code simulation. To fully unleash the potential of LLMs in logic code simulation, we introduce our method, Dual Chains of Logic(DCoL). It will guide LLMs to adopt a dual-path thinking approach, helping LLMs navigate through reasoning traps effectively and reducing the confusion of LLMs during logic solving. Finally, we further explore the effectiveness, strengths, and pitfalls of LLM-based logic solvers.

We highlight our contributions as follows:
\begin{itemize}
\item We are the \textbf{first} to propose a novel task, logic code simulation, for accessing LLMs' capability of directly solving logic problems encoded in programs with model inference.
\item We collect three new datasets for the logic code simulation task. We also conduct comprehensive evaluations on different LLMs.
\item We propose a novel LLM-based code simulation technique, DCoL, resulting in a notable improvement in accuracy by 7.06\% with GPT-4-Turbo. 
\end{itemize}

\section{Background and Related Work}

\subsection{Large Language Model}
Large Language Models (LLMs) are generative models based on the pre-trained Transformer architecture. Most LLMs utilize a generative model architecture, where given a sentence of $n$ tokens, the model is trained to maximize the likelihood of the ground-truth token $t_i$ at the current time step $t$ based on its preceding sequence $t_{i-1},...,t_1$. The training of LLMs typically follows three main processes: unsupervised training on large amounts of unlabeled text data without explicit human annotations, supervised fine-tuning on labeled data relevant to specific tasks or domains, and reinforcement learning on feedback from human annotators or evaluators. Leveraging extensive multimodal data and employing pre-training and fine-tuning techniques, LLMs have demonstrated state-of-the-art performance across various downstream tasks, such as machine translation, numerical reasoning, and code clone detection, with minimal examples (few-shot) or task-specific prompt instructions (zero-shot).

\subsection{Logical Problems Solving with LLMs}
We begin by identifying logical problems to be addressed in this work. Generally speaking, the hardest logical problem underlies the framework of satisfiability modulo theories (SMT). SMT is a variation of the SAT problem for first-order logic (FOL), with the interpretation of symbols constrained by specific theories, such as real arithmetic and bit vectors. Our study encompasses several subsets of SMT problems, including propositional logic, SMT problems themselves, and a segment of first-order logic within the solving ability of SMT, along with external theories defining constraints. Thus, we focus on SMT-based languages and solvers, to universally encode and solve those logical problems.

We elaborate on several samples to be solved in Fig \ref{fig2}.  In the realm of propositional logic, we incorporate variables like $x$ to enhance conceptual clarity. The proposed hypothesis, \textit{Reggae(x) → ExpressFeelings(x)}, can be derived with deductive reasoning. We transform the judgment process into a satisfiability dilemma by treating the assumption as a constraint. FOL supports quantifiers $\forall$ and $\exists$ on the basis of propositional logic, introducing uncertainty into problems. In this case, the assumption remains indeterminate. We check both the affirmation and negation hypotheses with SMT solvers, to align with the True/False/Uncertain output form. The uncertain cases are equivalent to UNSAT for both affirmation and negation hypotheses.
Concerning SMT-based problems, theories also contribute to their complexity. In the SMT-BitVec case, the target is to ascertain whether the Left-Hand-Side (LHS) equals the Right-Hand-Side (RHS) for a bit-operation-oriented issue. The true/false corresponds to SAT/UNSAT. Conversely, the SMT-Real-Arthmetic is drawn from the Z3py manual. The Z3 solver produces an unknown result as $x$ is involved in the exponentiation of $2^x$, rendering this problem non-polynomial. In contrast, individuals can readily find a solution as $x = \log_2 3$.

Despite directly solving reasoning problems in natural language~\cite{wei2022chain, zhou2023leasttomost, lyu2023faithful, ling2024deductive}, several researches are interested in symbolic-based logic solving triggered by the improvement in code generation ability of LLMs. These works are implemented within a two-step paradigm: initially translating language-based reasoning tasks into codes suitable for logical solvers, followed by calling external solvers to execute generated codes. To address potential issues like syntax errors and missing elements in the codes, Logic-LM~\cite{pan2023logic} iteratively refines generated logical form with error messages, while LINC~\cite{olausson2023linc} utilizes a voting mechanism to filter out errors and provide robust results. Additionally, SATLM~\cite{ye2024satlm} parsing all problems into FOL formulas and solving with Z3 solver. Besides, LoGiPT~\cite{feng2023language} is a fine-tuned LLM aiming for deductive reasoning, whereas SymBa~\cite{lee2024symba} leverages backward chaining to conduct logic reasoning reversely. SoLA~\cite{zhang2024sola} combines LLMs and differential solver layers to address SAT problems.

Our work differs from prior works in various aspects. First, The source inputs are not limited to natural language questions, but also logic codes generated from the software testing or verification procedure in a real-world setting. Second, the key factor in our research is how LLMs directly comprehend and simulate the code rather than on code translation and generation from natural language. Third, we aim to provide extensive evaluations of various problem types instead of one specific genre of questions.

\subsection{Code Simulation with LLMs}
Code simulation is a relatively new topic of interest. It is defined as predicting the concrete outputs of codes aligning with the execution results. Recent studies on Tuning-completeness of transformers and LLMs~\cite{perez2021attention, wei2022statistically, giannou2023looped, schuurmans2023memory} have suggested that LLMs are capable of interpreting instructions from code, showing potential to simulate the execution of code and algorithms as analog models. The previous works encourage LLMs to simulate a variety of tasks, such as keyboard and mouse actions on computers~\cite{kim2024language} and optimizers~\cite{yang2024large}. While transformer-based models are trained to predict the execution traces of codes~\cite{liu2023code} and UNSAT cores~\cite{shi2023satformer}, a recent attempt at simulating code with LLMs has been proposed~\cite{la2024code}. Their results indicate that LLMs can execute instructions sequentially. When facing long programs and complex procedures, several tackles such as memorization. The setting of our proposed logic code simulation focuses on simulating specialized codes of logical solvers instead of codes for general purposes. The potential outcome space is limited to either SAT or UNSAT, making it more favorable for obtaining accurate predictions since logical codes can be more easily expressed and comprehended in natural language.

\section{Method}
In this section, we first introduce related LLMs and the data-collecting procedure. Afterward, we present a new prompt technique, \textit{Dual Chains of Logic} (DCoL), aiming to improve the accuracy, reasoning process, and robustness of the logic code simulation task.

\subsection{Involved LLMs}
The objective of this study is to assess LLMs' reliability and limitations in simulating various forms of logic codes compared to conventional logical solvers like Z3. To accomplish this, we employ both open-source LLMs such as the LLaMA family~\cite{touvron2023llama, roziere2023code}, and close-source but strong LLMs such as the GPT family as our base models. We present details of these models as follows:

\textbf{GPT-3.5 Turbo}: GPT-3.5 Turbo, also known as ChatGPT, is a decoder-only network based on the Transformer architecture, comprising 175 billion parameters. It has undergone pretraining on 45 terabytes of text data gathered from diverse sources like books, articles, and websites. Moreover, GPT-3.5 Turbo is optimized for invoking its capability with Reinforcement Learning from Human Feedback (RLHF) technique. The GPT-3.5 Turbo-0125 supports a context window of 16,385 tokens.

\textbf{GPT-4 Turbo.} GPT-4 Turbo is an advancement of GPT-4, offering increased capabilities with an updated knowledge cutoff extended to April 2023. It also features a 128k context window, equivalent to 300 pages of text in a single prompt.

\textbf{LLaMA-2.} LLaMA-2, an open-source foundational model introduced by Meta Platforms, Inc., has been trained on 2 trillion tokens and is designed to accommodate a context length of 4096 by default. LLaMA-2 models have been refined with the input of more than 1 million human annotations, specifically tailored for chat purposes. LLaMA-2 offers four model sizes: 7B, 13B, 34B, and 70B. This study focuses on the evaluation of the 13B model.

\textbf{Code LLaMA.} Code LLaMA is a refined version of LLaMA-2 designed to support various code-related activities like coding, testing, explaining, and finishing code segments. We also deploy CodeLLaMA-13B for our evaluation.

\subsection{Datasets}
In contrast to prior research focusing on natural language-based issues, our study aims to investigate the capability of Large Language Models (LLMs) in emulating logic code execution. Some existing datasets are unsuitable for our research due to challenges and inaccuracies encountered during the code translation process. Translation challenges in datasets like FOLIO and AR-LSAT have been only partially resolved, with successful translation rates reaching 66.7\% and 21.8\%~\cite{saparov2023language}, respectively. Therefore, we have opted for ProntoQA, a synthetic propositional-logic QA dataset that can be translated into logic form nearly perfectly, providing a benchmark for comparison with translation-centric approaches. Nevertheless, it is notable that the execution accuracy by GPT-4 of ProntoQA reported is 83.2\%, suggesting the persistence of translation inaccuracies. We apply the hardest 5-hop subset selected in LogicLM~\cite{pan2023logic}. The questions such as `True or false: Alex is not shy.' are encoded into solvers by treating it as a constraint `IsShy(Alex)', then the SAT/UNSAT is mapped to True/False.

\begin{table}[!ht]
\centering
\caption{Dataset information provided. Format means the input format of questions. Z3Py is based on Python, while SMTLIB is the standard language of SMT solvers. Mean LOC represents the average Lines of Code (LoC) of examples.}
\label{tab1}
\begin{tabular}{c|c|c|c|c}
\hline
\multirow{2}{*}{\textbf{Dataset}} & \multirow{2}{*}{\textbf{Formulation}} & \multirow{2}{*}{\textbf{Format}} & \multirow{2}{*}{\textbf{\# of Samples}} & \multirow{2}{*}{\textbf{Mean LoC}} \\
 & & & &\\
\hline
ProntoQA & PL & NL & 500 & - \\

Z3Tutorial & SMT & Z3Py & 37 & 9.90 \\

Z3Test & SMT & Z3Py & 85 & 8.37 \\

SMTSIM & SMT & SMTLIB & 104 & 14.36\\
\hline
\end{tabular}
\end{table}

To delve deeper into simulating real-life logical codes, we propose 3 new datasets from various sources. (1) We gathered 31 code examples from the Programming Z3 tutorial~\cite{Bjorner2019}, culminating in a new dataset named \textit{Z3Tutorial}. (2) Additionally, the Z3 official repository, Z3test\footnote{https://github.com/Z3Prover/z3test}, offers a range of test samples for the Z3 solver, from which we extracted 167 Python cases to form the  \textit{Z3Test} dataset. These cases were then categorized into three question types: logic-only, arithmetic, and type-inference. We discard the type-inference questions such as \textit{is\_add(x+y)} in our primary experimental setting, but will be introduced as a challenge for logic code simulation. (3) we included several codes in SMT-LIB format from the SMT-COMP 2023~\cite{beyer2023competition}. These instances are predominantly derived from industrial software or established logical problem collections. We have chosen various external theories to enhance the intricacy of the problems, including real arithmetic (both linear and non-linear), bit vectors, uninterpreted functions, and strings. Our \textit{SMTSim} dataset consists of 102 samples in total. For further insights, we present a comprehensive overview of our curated dataset in Table \ref{tab1}. Note that the complexity of a logic problem is determined by both the quantity of constraints and the difficulty of logic expressions. The number of lines is only a partial factor of the overall complexity. All those code snippets are ensured to only produce the result of SAT or UNSAT.

% The workflow of our method is prompt-based. For each model, we first deploy strong baselines such as Chain of Thought (COT) prompting~\cite{wei2022chain} and Plan-and-Solve Prompting~\cite{wang2023plan} to discover how LLMs reason logic problems with different 

\subsection{Prompting Baselines}
The baseline Prompting methods considered in this work are listed below:
\begin{itemize}
    \item Standard Prompting (SD)~\cite{brown2020language}. SD straightly asks LLMs to predict the execution results of logic codes. The LLMs will return only SAT/UNSAT, sometimes with short explanations.
    \item Chain-of-Thought (COT)~\cite{wei2022chain}. COT encourages LLMs to generate a systematic solution by explaining each step before presenting the final answer. The LLMs will return a reasoning path tackling the issue raised.
    \item Plan-and-Solve Prompting (PS)~\cite{wang2023plan}. PS instructs LLMs to devise a detailed plan for addressing issues and solve plan targets 
    sequentially. 
    \item Chain of Simulation (CoSm)~\cite{la2024code}. The CoSm method enhances the conventional Chain of Thought prompting technique by circumventing the drawbacks associated with memorization in code simulation tasks. It aims to track the program trace and forces LLMs to simulate each instruction sequentially.
\end{itemize}

\begin{figure*}[!ht]
\centering
\includegraphics[width=18cm]{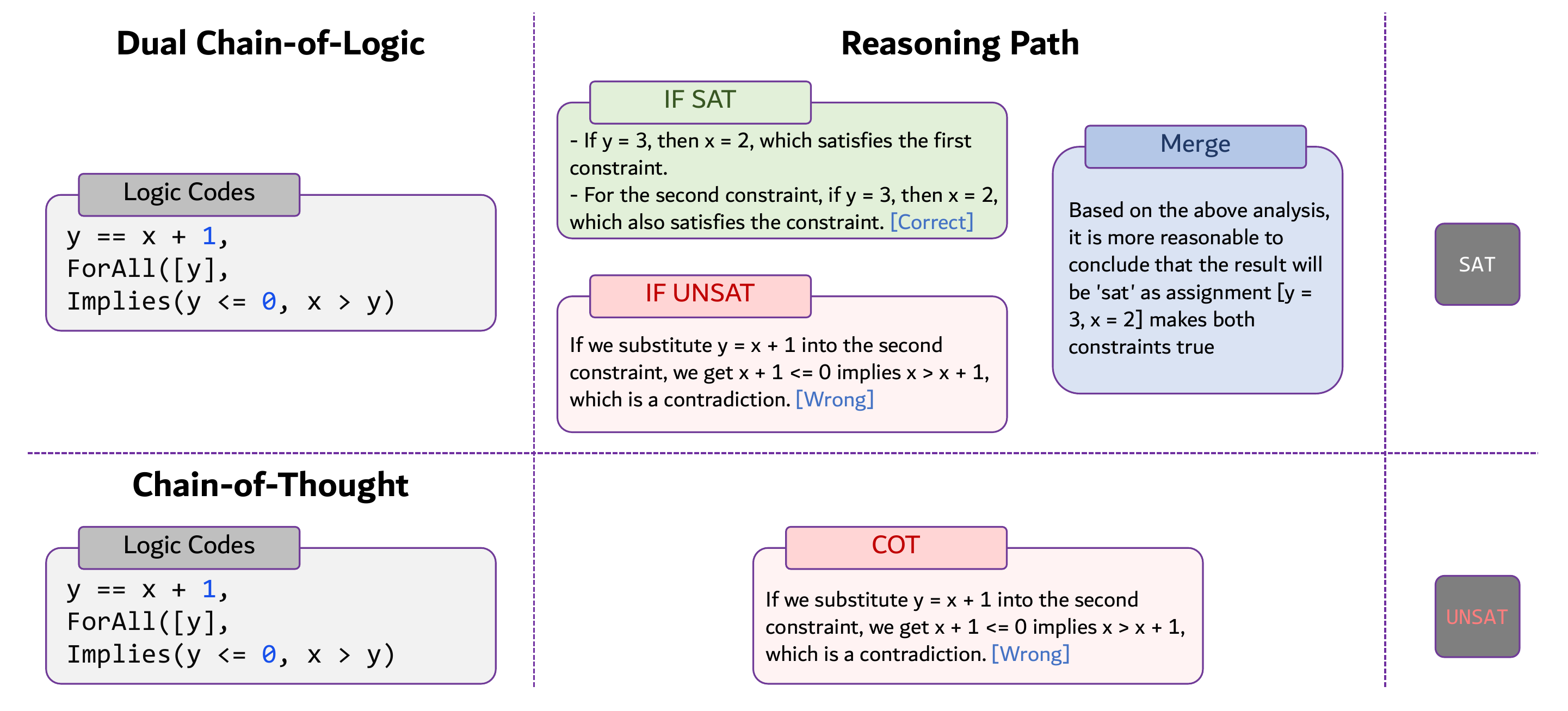}
\caption{Overview of the DCoL method: DCoL offers two hypotheses, SAT (satisfiable) and UNSAT (unsatisfiable), for logic code simulation. The LLMs verify these hypotheses individually before combining them to reach the final decision, while the COT method only outputs one possible reasoning path.}
\label{fig3}
\end{figure*}

Besides, Self-Consistency (SC)~\cite{wang2022self} employs a majority voting mechanism on different reasoning paths for robust prediction. The insight behind this method is maintaining a consistent As SC is a parallel prompting method to those above, it can be applied in conjunction with prompting methods including ours. Thus, we evaluate it as a stand-alone plugin.

Note that our proposed task is a novel one focusing on codes as initial input. Thus, methods for processing natural language problems~\cite{pan2023logic, ye2024satlm, zhang2024sola}, are no longer compatible with our framework. Furthermore, these approaches are limited in their ability to address a specific portion of the task we have outlined, such as SAT~\cite{zhang2024sola} and SMT-Int (SMT questions with integer computing theory)~\cite{pan2023logic, ye2024satlm}. As a result, we do not include them as baselines in our study.

A notable observation is that the typical few-shot prompt, which is also known as in-context learning is ineffective in our experimental setting, which is also reported in similar tasks such as code simulation~\cite{la2024code} and discussed in several works~\cite{min-etal-2022-rethinking}. Few-shot prompt learning may not work on a task whose input-label correspondence is not already captured in the LLMs. In our logic code simulation task, elementary operations in different subtasks such as different function calls and various theories can be challenging for in-context learning. Besides, considering the intricate nature of logic problems, the templates heavily exceed the LLMs' context limit. Thus, we focus on the zero-shot prompting scheme in this research.

\subsection{DCoL: Dual Chains of Logic}
\begin{figure}[ht]
\centering
\includegraphics[width=9cm]{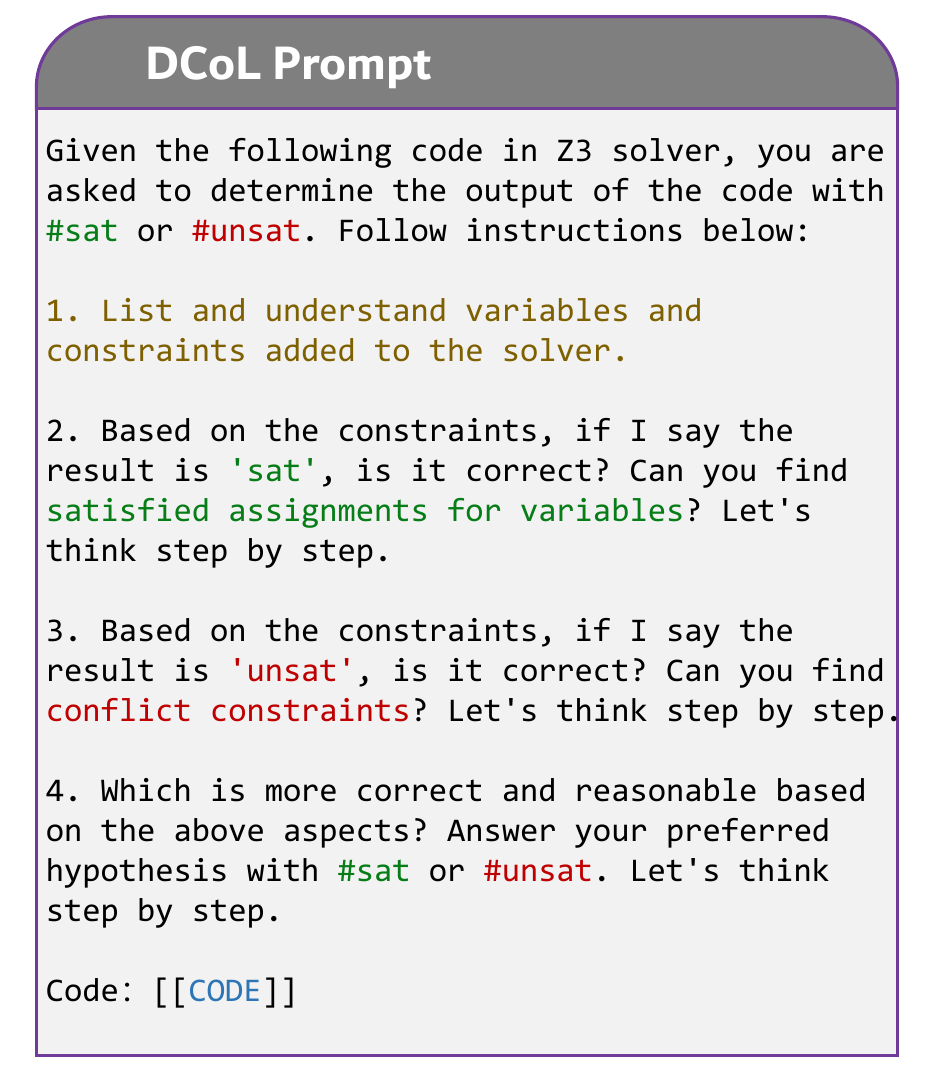}
 \caption{Prompting template of the DCoL method. Prompts can be modified slightly according to specific tasks.}
\label{fig4}
\end{figure}

Among these approaches, the concept of reasoning paths is crucial, as they lead to more precise outcomes. Nevertheless, when dealing with general questions like mathematical computations and text generation, the answer search space is practically infinite, posing challenges in constructing reasoning paths. In contrast, logic problems that can be converted into solvers typically result in \textit{dual} possible outcomes: SAT or UNSAT. The reasoning paths of humans for determining whether a set of constraints is SAT or UNSAT are diverse:
\begin{itemize}

\item[(1)] If the result is SAT, we should find a legal assignment of variables to verify it. Consider the case presented in Figure \ref{fig4}. Assuming that the given codes produce the output SAT, an arrangement of variables that fulfills all constraints serves as compelling evidence. One possible assignment is $x = 2, y = 3$, which satisfies the first constraint $y \equiv x + 1$. In regard to the second constraint, $\forall y\ [(y \le 0) \implies (x < y)]$, $y$ is a \textbf{independent variable}\footnote{Independent variable $y$ can be replaced with any symbol, such as $\forall z\ [(z \le 0) \implies (x < z)]$}, unrelated to the $y$ defined in $y \equiv x + 1$. Therefore, only $x = 3$ is relevant in this context, and this assignment meets the second constraint.
\item[(2)] If the result is UNSAT, we should find a set of minimal constraints where conflicts always exist, which is also known as UNSAT core. Replacing $x > y$ with $x < y$ in the second constraint transforms the problem into an UNSAT one. The UNSAT core solely consists of the second constraint because, regardless of the value assigned to $x$, there exists a value for $y$ that is non-positive and less than the given $x$ value. Such that the second constraint does not always hold.
\end{itemize}

% 这里需要有一段承上启下的解释段，从HUMAN到LLM

However, LLMs cannot fully grasp the dual implicit reasoning chains for logic problems during one execution time. Among the dual paths, at least one leads to an accurate solution. Thus, by guiding LLMs to reason via both SAT/UNSAT paths, LLMs can produce different reasoning paths for contrary goals, and generate promising answers with explanations.

To this end, we propose a novel prompt method, Dual Chains of Logic (DCoL). Fig. \ref{fig3} illustrates our proposed DCoL approach. DCoL encourages LLMs to discard incorrect wrong prediction UNSAT after integrating dual paths of thinking, while COT only generates a wrong reasoning path, leading to the prediction error. As illustrated in the template of DCoL from Fig. \ref{fig4}, the initial step involves instructing LLMs to extract variables and constraints for reference. Subsequently, we present the dual assumptions individually and task LLMs with solving them for different targets. The COT technique is leveraged in this step to provide more specific explanations. At last, LLMs combine these chains and opt for the more precise and rational one along with corresponding explanations as their final prediction.

DCoL, in contrast to COT, steers clear of falling into the erroneous reasoning trap illustrated in Fig. \ref{fig3}. Since $y$ functions as an independent variable, it is inaccurate to interchangeably use $y=x+1$ in the second constraint, given the realm that $y$ in both constraints does not hold the same significance. This error, albeit subtle, often goes unnoticed even by humans lacking specialized knowledge. However, DCoL consistently provides a precise reasoning approach for obtaining accurate outcomes by thinking on the dual side of logic problems.

\begin{table*}[htbp]
    \centering
   \normalsize
    \caption{Performance of LLMs with different prompt techniques. The \textbf{bolded} numbers denote the best performance in this dataset, while results in \cellcolor{gray!25}grey boxes represent the best result among all the prompt methods. All experiments are conducted three times.}
    \begin{tabular}{cccccccccc}
    \toprule
    \multicolumn{2}{c}{\multirow{2}{*}{\textbf{Method}}} & \multicolumn{3}{c}{\textbf{Z3Tutorial}} & \multicolumn{3}{c}{\textbf{Z3Test}} \\
    \cmidrule(lr){3-5}\cmidrule(lr){6-8}\cmidrule(l){9-10}
    & & \textbf{Accuracy} & \textbf{Unknown} & \textbf{Exe. Acc.} & \textbf{Accuracy} & \textbf{Unknown} & \textbf{Exe. Acc.} \\
    \midrule
   \multicolumn{2}{c} {Z3} & 100.0 & - & 100.0 & 98.80 & 1.20 & 100.0 \\ % New row
    \midrule
    \multirow{5}{*}{\begin{tabular}[c]{@{}c@{}}GPT-3.5 \\Turbo\end{tabular}} 
    & SD & 67.74 & - & 67.74 & 70.59 & - & 70.59 \\
    & PS & 67.74 & - & 67.74 & 72.54 & 3.52 & 75.19 \\
    & CoSm & 63.44 & - & 63.44 & 71.76 & - & 71.76 \\
    & COT & 68.82 & - & 68.82 & 74.11 & 0.39 & 74.40 \\
    & DCoL & \cellcolor{gray!25}70.97 &  & 70.97 & \cellcolor{gray!25}74.90 & - & 74.90 \\
    \midrule
    \multirow{5}{*}{\begin{tabular}[c]{@{}c@{}}GPT-4 \\ Turbo\end{tabular}} 
    & SD & 84.95 & - & 84.95 & 76.47 & 3.53 & 79.27 \\
    & PS & \cellcolor{gray!25}\textbf{86.02} & - & 86.02 & 82.35 & - & 82.35 \\
    & CoSm & 81.72 & - & 81.72 & 80.39 & 2.35 & 82.32 \\
    & COT & \cellcolor{gray!25}\textbf{86.02} & - & 86.02 & 81.18 & 2.35 & 83.13 \\
    & DCoL & \cellcolor{gray!25}\textbf{86.02} & - & 86.02 & \cellcolor{gray!25}\textbf{83.53} & 1.76 & 84.52 \\
    \midrule
    \multirow{5}{*}{\begin{tabular}[c]{@{}c@{}}LLaMA\\2-13B\end{tabular}} 
    & SD & 35.48 & 36.55 & 55.91 & 49.90 & 17.76 & 58.78 \\
    & PS & 44.09 & 36.55 & 69.49 & 51.69 & 29.54 & 66.46 \\
    & CoSm & 32.25 & 37.63 & 51.71 & 47.70 & 31.33 & 63.37 \\
    & COT & 45.16 & 35.48 & 70.00 & 44.11 & 37.92 & 63.07 \\
    & DCoL & \cellcolor{gray!25}51.61 & 34.40 & 78.67 & \cellcolor{gray!25}54.69 & 24.95 & 67.16 \\
    \midrule
    \multirow{5}{*}{\begin{tabular}[c]{@{}c@{}}Code\\LLaMA\\13B\end{tabular}} 
    & SD & 44.08 & 33.33 & 60.75 & 44.31 & 28.00 & 61.07 \\
    & PS & 46.23 & 43.01 & 67.74 & 45.90 & 42.71 & 67.26 \\
    & CoSm & 43.01 & 36.55 & 61.29 & 46.50 & 39.12 & 66.06 \\
    & COT & 39.78 & 47.31 & 63.44 & 41.71 & 48.10 & 65.76 \\
    & DCoL & \cellcolor{gray!25}52.68 & 34.40 & 69.89 & \cellcolor{gray!25}49.30 & 36.32 & 67.46 \\
    % \midrule
    % \multicolumn{2}{c}{\textcolor[rgb]{ .373,  .616,  .984}{Avg impv}} 
    % & - & - & - & - & - & - \\
    \bottomrule
    \end{tabular}%
    \label{tab2}
\end{table*}

\section{Experimental Results}
In this section, we will revisit the research questions raised above and examine them by breaking them down into more manageable parts.

\subsection{Effectiveness on Logic Code Simulation (\textbf{RQ1})}
We employ four LLMs, GPT-3.5-Turbo, GPT-4-Turbo, LLaMA-2-13B, and CodeLLaMA-13B, with different prompting methods, to explore their capability of them on simulating logic codes. The results are provided in Table ~\ref{tab2}. Two datasets based on Z3Py are utilized for a comprehensive analysis. 

We utilize two additional metrics, namely "unknown rate" and "execution accuracy" to further investigate the effectiveness of logic code simulation. 

\textbf{Unknown rate} indicates the percentage of instances such as `If a solution exists, the output will be SAT; otherwise UNSAT', where LLMs struggle to provide a definitive answer. An unknown output can also be caused by errors in the program, such as `The code should output ERROR rather than SAT or UNSAT'.

\textbf{Execution accuracy} refers to the precision among instances that LLMs confirm to be SAT/UNSAT. This measure aids in assessing the reliability of the outputs generated by LLMs, The execution accuracy $EA$ is calculated with:  

\begin{equation}
    EA = \frac{Acc}{1-UNK}
\end{equation}

A model can have low accuracy while obtaining a high execution accuracy. If we apply a random guessing procedure on those unknown examples, the accuracy will be acceptable then.
Note that all methods obtain execution accuracy greater than 50\%, which means LLMs can perform at least stronger than random guessing. 

The main observations are summarized below:

\textit{GPT can pretend logic solvers. } As shown in Table \ref{tab2}, GPT shows its capability to simulate logic codes even with the standard prompting method (SD).
In the Z3Tutorial dataset analysis, it is evident that employing standard prompts can yield an accuracy exceeding 80\% when utilizing GPT-4 Turbo, known for its high resilience among LLM options. Even when employing SD prompts with GPT-3.5, an accuracy of 67.74\% can be attained. As for the Z3Test dataset, GPT-3.5 achieves over 70\% accuracy, while GPT-4 successfully solved 80\% logic problems in this dataset. This phenomenon comes from the combination of code understanding and logical reasoning for GPT models. 

\textit{LLaMA is not confident to make predictions. } In contrast to GPT models, LLaMA models often produce more 'unknown' outcomes, despite having comparable or better performance accuracy. In the case of the Z3Tutorial dataset, approximately one-third of the predictions result in unknown outcomes. This occurrence may stem from the LLaMA models' limitation in comprehensively grasping the code, leading them to address relatively straightforward questions in the dataset. Furthermore, our research reveals that LLaMA exhibits weaknesses in adhering to instructions, yielding challenging responses to interpret and analyze. Therefore, we prioritize GPT families to convey our research better.

\textit{COT is a strong baseline. } 
We conducted a thorough study of different prompting techniques to enhance logical code simulation. Among these methods, Chain-of-Thoughts (COT) serves as a strong baseline. Table \ref{tab2} shows that COT boosts standard prompting with 1.18\% on the Z3Tutorial dataset and 3.92\% on the Z3Test dataset. The superior performance on COT demonstrates the effectiveness of the reasoning path.

\textit{Think on logic rather than think on code. } 
A representative method performing `Think on Code' is Chain of Simulation (CoSm), which forces LLMs to simulate the program instruction by instruction. However, this method is only suitable for simulating problems with sequential execution paths. Logic reasoning is a more complex issue as the execution chain of logic codes is hidden under external solver libraries such as the Conflict Driven Clause Learning algorithm applied in Z3 solver. To solve logic problems, LLMs should pay more attention to logical reasoning to derive the correct execution path. Nevertheless, the execution of code should not be ignored in our task. A typical error case involves adding constraints to a solver that are never actually verified, where LLMs should disregard these constraints.

\textit{DCoL prompt is effective. }
Our proposed prompt method, Dual Chains of Logic, offers LLMs the idea of trying two separate ways of problem-shooting. We observe that among all LLMs, DCoL can steadily enhance the performance of logic code simulation across all datasets. The baseline COT is less competitive on GPT models than DCoL. However, when utilizing LLaMA models, our proposed DCoL significantly outperforms all baselines. The explanation lies in behind the phenomenon can be that with the evolving of LLMs, they also acquire the ability for implicitly multi-way reasoning. Furthermore, GPT models have shown a tendency to deviate from our guidelines in experiments by producing a single reasoning process directly, resulting in less impact of our proposed DCoL prompting while reasoning.

\begin{figure}[ht]
\centering
\includegraphics[width=8.5cm]{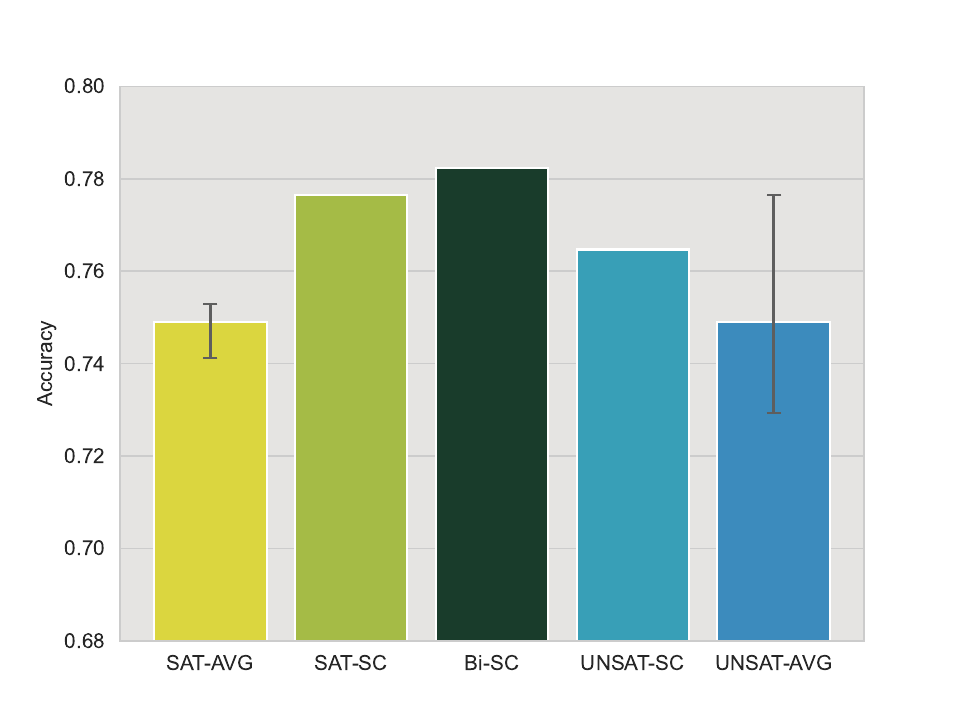}
 \caption{Prompting template of the DCoL method. Prompts can be modified slightly according to specific tasks.}
\label{fig5}
\end{figure}

\textit{Bi-directional Self-Consistency improves performance. }
A recent study suggests that the sequence in which premises are given to LLMs significantly impacts their performance in natural language reasoning tasks~\cite{chen2024premise}.
The Dual Chains of Logic prompt we introduced will present SAT and UNSAT questions to LLMs in a successive manner.  By swapping the order of SAT/UNSAT reasoning path in DCoL, LLMs can show different tending for the final prediction. Inspired by the idea of self-consistency, To explore this further, a validation experiment was designed based on the concept of self-consistency. The experiment involved sending 3 SAT-first and 3 UNSAT-first DCoL prompts to Z3Test via GPT-3.5-Turbo separately, followed by aggregating the responses using a majority voting approach. The experimental results are depicted in Fig. \ref{fig5}, showing that the bi-directional Self-Consistency (Bi-SC) mechanism resulted in a 3.34\% improvement to the average performance by bridging the gap between execution orders.

\textbf{Answer to RQ1:} LLMs, especially GPT families, can effectively stimulate the reasoning results of logic solvers. The simulation accuracy can be further improved with our proposed DCoL prompt and Bi-directional Self-Consistency mechanism.

\subsection{What are the strengths of LLM-based logic code simulation? (\textbf{RQ2})}
Besides conducting comprehensive experiments on Z3-based programs, we further explore the promises of logic code simulation to answer this research question.

\textit{LLM can simulate generated logic codes.}
This part is motivated by the failure reported by Logic-LM~\cite{pan2023logic}. Logic-LM tackles logic inference challenges in natural language employing a translation-solver framework. While Logic-LM demonstrates high proficiency in generating symbolic codes for the ProntoQA dataset, achieving close to 100\% successful execution rate, its precision drops to 61\% when calling the symbolic solvers. The root cause should be \textbf{LLMs can generate unreliable codes}, leading to incorrect results or even worse cases. We point out that the consistency for code generation and code understanding can resolve this issue, by leveraging logic code simulation procedure on generated code. To this end, we conduct a comparative study on the basis of the failure of Logic-LM, as shown in Table \ref{tab3}. The approach of \textit{COT + CodeGen} regards code generation as an auxiliary in-context learning task, while \textit{NL + Code} involves providing the original language input along with the generated codes in the previous step. We observe that even with low-quality codes, direct simulation achieves a performance similar to that of standard prompting for natural language problems, whereas \textit{NL + Code} obtains a comparable accuracy of 75.80\% to \textit{COT + CodeGen} when provided contexts, surpassing the Logic-LM baseline that depends on an external solver.

\begin{table}[htbp]
  \centering
  \caption{ProntoQA Comparison}
  \label{tab3}
  \setlength{\tabcolsep}{15pt}
  \begin{tabular}{cc}
    \toprule
    \textbf{Method} & \textbf{Accuracy} \\
    \midrule
    \multicolumn{2}{c}{\textbf{Direct Solving}} \\
    \midrule
    SD & 51.20\\
    COT &  72.00 \\
    COT + CodeGen  & 76.80\\
        \midrule
    \multicolumn{2}{c}{\textbf{Tranlation + Solver}} \\
    \midrule
    Logic-LM &  61.00 \\
    \midrule
    \multicolumn{2}{c}{\textbf{Tranlation + Simulation}} \\
    \midrule
    Code Only & 50.20\\
    NL + Code & 75.80\\
    \bottomrule
  \end{tabular}
\end{table}

\textit{LLMs are robust simulators.}
We have already demonstrated the capability of Large Language Models (LLMs) in code simulation, indicating that LLMs can comprehend our intentions based on the provided code. Since LLMs can predict the outcome of code without actually executing it, they are more tolerant of syntax errors compared to logic solvers. To assess the robustness of code simulation in LLMs, we conducted an experiment involving four distinct syntax error patterns:
\begin{itemize}
    \item Mismatched Parentheses: there is an inconsistency in the placement of parentheses, such as an unequal number of opening and closing parentheses or improper nesting, such as $(a + b$ or $func(a, b))$.
    \item Misspelled Variable or Function Names: incorrect spelling or naming of variables or functions within the code, such as $x1$ and $xl$.
    \item Mixing Z3 Library Grammar: elements specific to the SMT-LIB, such as $(declare-fun\ \langle symbol\rangle\langle sort\rangle^{*})$, are incorrectly integrated into the code syntax.
    \item Mixing First-Order Logic Grammar: elements of first-order logic grammar, such as quantifiers $\forall$ and $\exists$, are incorrectly incorporated into the code syntax.
\end{itemize}

\begin{figure}[ht]
\centering
\includegraphics[width=9cm]{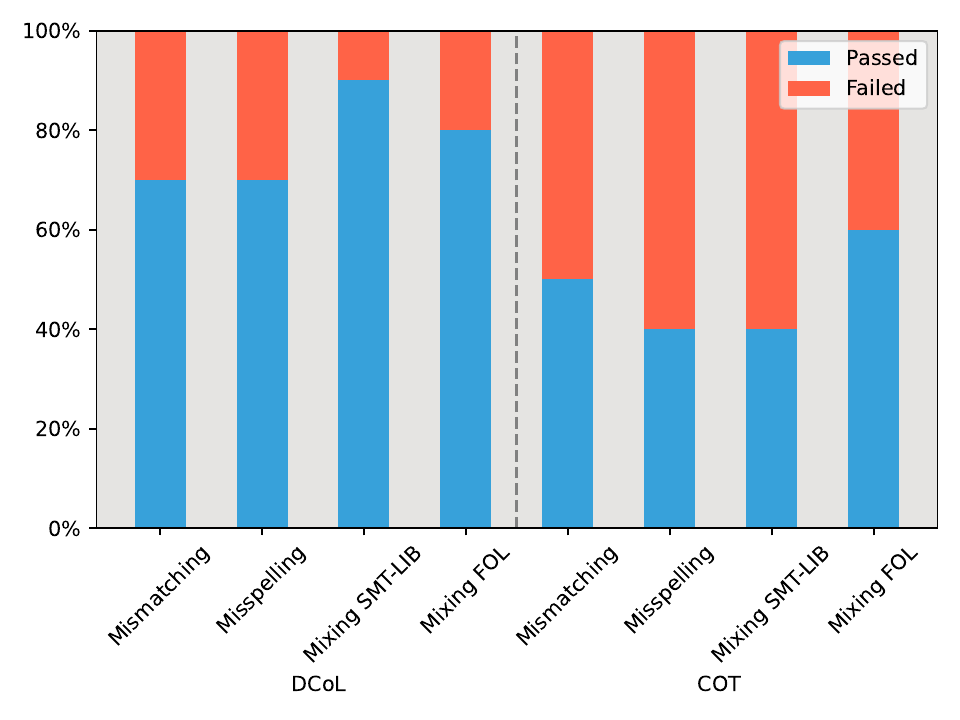}
 \caption{Performance drop against code-level modifications}
\label{fig6}
\end{figure}

For each syntax error pattern, we randomly selected 10 cases from the Z3 Test dataset to introduce errors. Subsequently, we applied both COT and our proposed method (DCoT) on the poisoned Z3 Test dataset. As illustrated in Fig. \ref{fig6}, the code simulation of LLMs with COT was significantly misled by the introduced syntax errors, leading to a decrease in accuracy of approximately 40\%. In contrast, syntax errors had minimal influence on the effectiveness of our method, which demonstrates \textbf{LLMs are robust simulators}.

\textit{LLMs exceed some theoretical limitations of solvers}
A motivated example of this strength is showcased in Fig. \ref{fig2}. Constraints like $2^x == 3$ go beyond the representation capability of SMT theories. Although it remains such a simple question for humans, typical solvers like Z3 output UNKNOWN in this case.

\begin{figure}[ht]
\centering
\includegraphics[width=8.5cm]{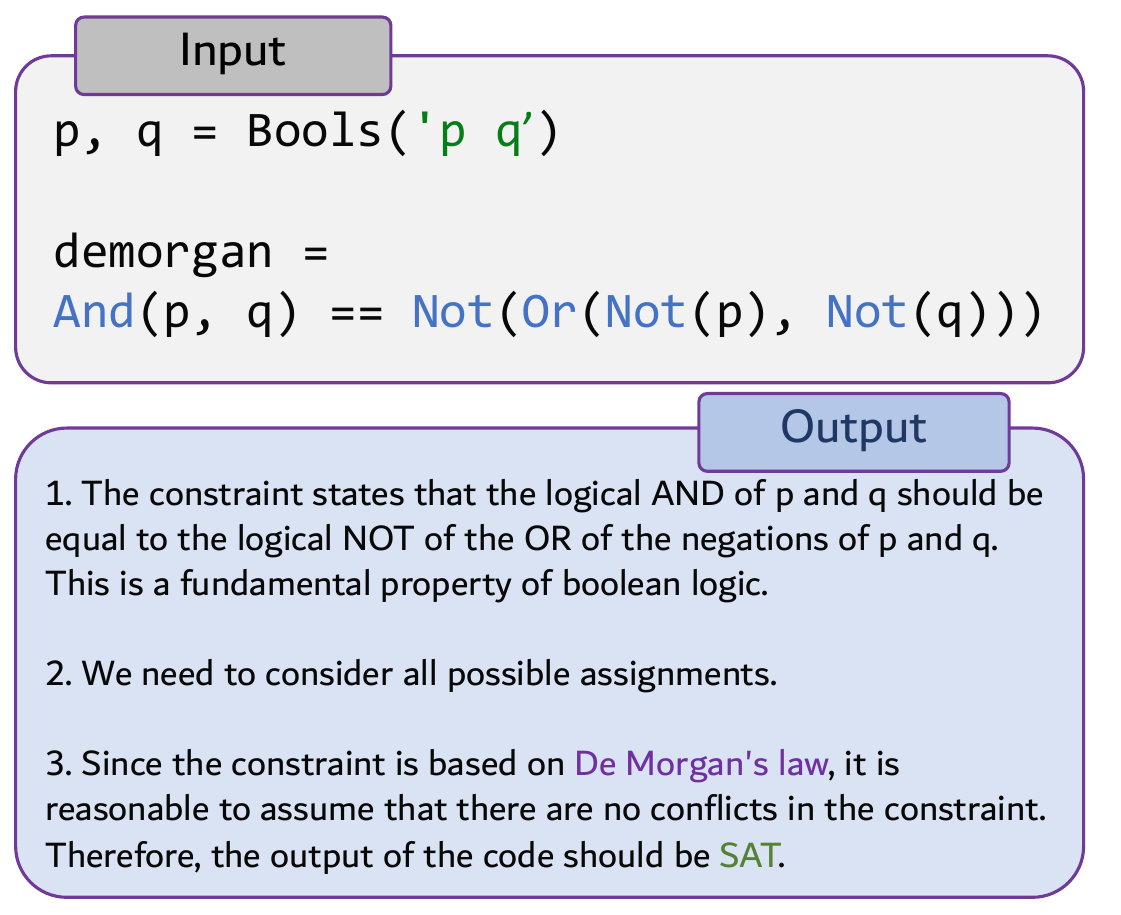}
 \caption{An example of De Morgan's Law}
\label{fign}
\end{figure}

\textit{LLMs leverage explainable knowledge for reasoning. }
We provide an interesting example in Fig. \ref{fign} targeting checking the satisfiability of De Morgan's Law. Despite enumerating all assignments for proving, LLMs build the connection from logic codes to De Morgan's Law and make the correct prediction with this external knowledge. We suggest that leveraging formulas and theorems is a good manual for complex problem-solving, and LLMs show their capability of utilizing knowledge for more explainable reasoning.

\textbf{Answer to RQ2:} LLMs demonstrate the capability to process a wider range of input types with a margin for error. Furthermore, they hold promise in addressing theories that logical solvers are unable to tackle.

% Another fact worth discussing is the heterogeneous computational models existing between LLMs and 

\subsection{RQ3: Pitfalls}
To address this research question, we conduct an error analysis to discover the mistakes in LLMs' behavior when facing code simulation challenges. Besides, we report the experimental result on a challenging dataset, \textit{SMTSim}, to investigate the ability of LLMs for solving

\textit{Error analysis. } We performed a comprehensive analysis of the findings derived from the Z3Test dataset utilizing three prompt methods: CoSm, COT, and DCoL, to discover the error types that occur in different reasoning paths. Our examination identified six categories of errors present in the unsuccessful results:
\begin{itemize}
    \item Inferring error: inferring error pertains to errors in causal deduction. For instance, if variables $a$ and $b$ are both true, but equations $a \wedge b$ are deduced as false, this constitutes an inferring error. Another example is when the reasoning process leads to contradictory conclusions, such as inferring $a>10$ initially but later treating it as $a<10$ in subsequent references, leading to misjudgment of results.
    \item Misunderstanding Satisfiability (SAT) error: occasionally, LLM accurately identifies a solution set for a problem expected to yield a satisfiable (SAT) result but instead returns unsatisfiable (UNSAT). This misunderstanding arises when the LLM erroneously treats an SAT problem as if it requires further solving, mistakenly assuming the presence of multiple solutions and consequently misclassifying it as UNSAT, leading to a misjudgment of the problem.
    \item Partial UNSAT error: In some instances, when the ground truth of SAT is considered, LLM may detect conflicting assignments while disregarding certain existing ones, leading to the classification of the problem as unsatisfiable. For instance, the formula $\{x=1, x>y\}$ is deemed satisfiable because a suitable set of assignments like $\{x=1, y=0\}$ can fulfill all the criteria. However, LLMs may occasionally identify assignments such as $\{x=1, y=2\}$ which conflict with the mentioned formula. In such cases, these assignments are designated as a UNSAT core by LLMs and consequently marked as UNSAT.
    \item Bit-vector arithmetic error: Issues have been identified in managing bitwise operations on bit vectors, particularly with certain operations posing challenges for the LLM. For example, when attempting to shift 3 (binary 11) to the left in a bit vector, the LLM might either provide an accurate outcome or misinterpret it as overflow, resulting in incorrect output.
    \item Real arithmetic error: This category pertains to computational mistakes involving real numbers. It differs from the abstract division in SMT-LIB by encompassing errors in both integer and floating-point computations. For instance, it addresses issues such as establishing the equality of positive and negative zeros in floating-point numbers, where the response from the software may lack clarity and may not guarantee the intended result.
    \item Commonsense error: LLM sometimes commits basic mathematical mistakes, like mistakenly equating 1 with 2 during calculations, or not identifying well-known library functions by their names, potentially affecting their intended use. This classification also encompasses situations where LLM misinterprets formulas or code, or struggles to grasp them fully.
\end{itemize}

\begin{figure}[ht]
\centering
\includegraphics[width=10cm]{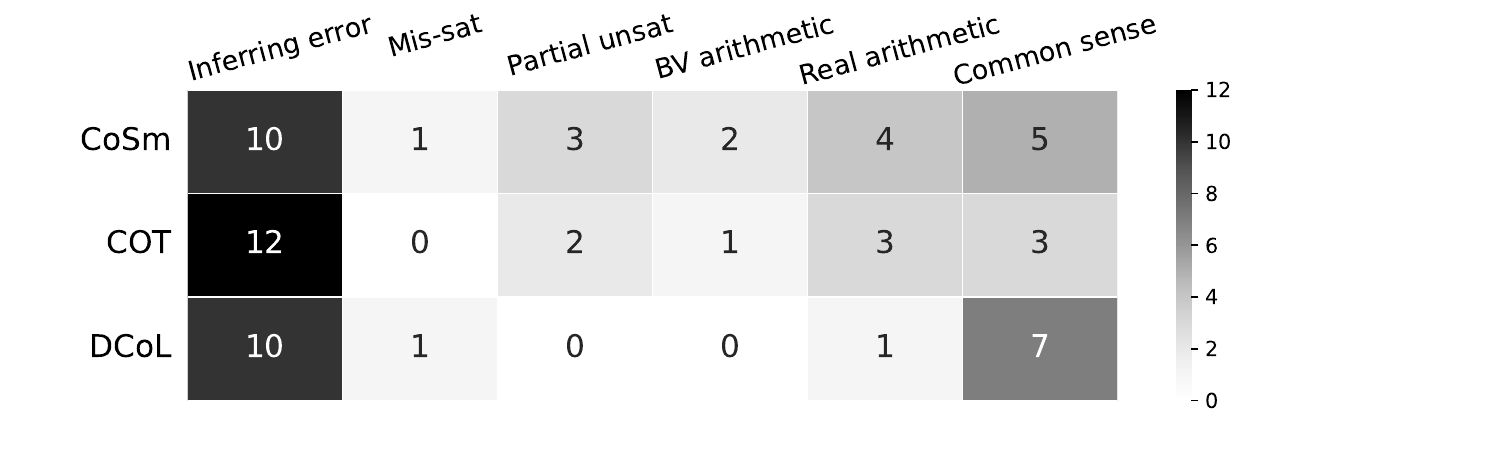}
 \caption{Error analysis among different methods}
\label{fig7}
\end{figure}

From our analysis, it is evident that our method, DCoL, aids the LLM in understanding satisfiability problems, thereby reducing the occurrence of misinterpretation of sat problems as equation-solving problems, and subsequently decreasing the incidence of misjudgment of problems. Moreover, in arithmetic problems, DCoL assists the large model in decomposing the initial arithmetic steps, resulting in significant performance improvements.

It's noteworthy to mention certain examples that, although not within the purview of our research, are prevalent within the industry but may not fare well with LLMs. One such example is the type judgment commonly utilized in Z3Py, exemplified by the function \verb|is_add()| as we will mention later. In Z3 Python code, \verb|is_add()| discerns whether the expression within parentheses denotes an addition operation. For instance, \verb|is_add(a + b)| is expected to return true, while \verb|is_add(a)| should return false. Because this function type isn't directly linked to logic but rather to language characteristics, it lies beyond the scope of our research. Nonetheless, focusing solely on such functions, it is interesting that this kind of example may lead to suboptimal performance for LLMs. For instance, \verb|is_add(1 + 2)| is anticipated to return false because $1 + 2$ is interpreted as a constant number in Z3Py, which may not only confuse LLMs but also confuse many people who are not familiar with language features.

Another type of example originates from the \textit{SMTSim} dataset collected from SMT-COMP 2023. Some SMT files are generated by intricate backend libraries, resulting in a significant number of variables and formulas within these examples. However, considering the token limit for each query of LLMs, we had to filter SMT files, removing examples surpassing a certain threshold of code lines or tokens per line. Nonetheless, many of these filtered examples contain queries that cannot be resolved by Z3 or other solvers, yielding unknown results. These instances hold potential value for research in this domain, and we aspire to delve deeper into this issue in future investigations.

\textit{Hard problems. }
\textit{SMTSim} dataset contains complicated logic code, with more lines of codes, variables, and theories. We regard it as a challenge for logic code simulation. We report the performance of GPT families in Table \ref{tab4}.

\begin{table}[htbp]
    \centering
    \caption{SMTSim Results}
    \begin{tabular}{ccccccc}
    \toprule
    \multicolumn{2}{c}{Method} & ACC.   & UNK. & Exe. Acc  \\
    \midrule
    \multicolumn{2}{c}{Z3 Solver} & 97.06  & 2.94 & 100.0 \\
    \midrule
    \multirow{3}{*}{\begin{tabular}[c]{@{}c@{}}GPT-3.5 \\Turbo\end{tabular}} 
    & COT & 5.88  & 85.29 & 33.33   \\
          & CoSm & 14.71 & 78.43 & 68.20 \\
          & DCoL & \textbf{54.9} & 6.86 & 58.94 \\
    \midrule
    \multirow{3}{*}{\begin{tabular}[c]{@{}c@{}}GPT-4 \\ Turbo \end{tabular}} 
    & COT & 51.96  & 19.61  & 65.92   \\
          & CoSm & 50.98 & 23.53 & 66.67 \\
          & DCoL & \textbf{58.82} & 5.82 & 62.45 \\
    \midrule
    \end{tabular}%
    
    \label{tab4}
\end{table}

We observe that COT and CoSm suffer from UNKNOWN prediction, while our proposed DCoL achieves over 50\% accuracy even with GPT-3.5-Turbo. However, the relatively low execution accuracy reveals that LLMs struggling to solve such complex datasets. This issue may stem from the abundance of variables and constraints present. In the SMTSim dataset, numerous instances occur infrequently but are deeply intertwined. Due to their limitations in handling extensive contexts, LLMs struggle to address complex logical compositions effectively.

\textbf{Answer to RQ3:} Although LLMs demonstrate satisfactory performance on several datasets, the potential drawbacks and limitations of LLMs for logic code simulation should not be underestimated.

\section{Conclusion and Future Work}
In this work, we introduce the novel concept of logic code simulation to evaluate the ability to comprehend, analyze, and simulate logic problems encoded in programs. We release three datasets collected from the solver community and perform an extensible assessment to evaluate the capability of logic code simulation for various LLMs. Furthermore, we propose DoCL, a novel prompt fashion, to improve the reasoning performance on code-based logic problems.

In the future, we are committed to further enhancing our DoCL, recognizing that there is considerable room for advancement. Leveraging the principles of DoCL, numerous techniques can be implemented to drive significant improvements. Moreover, we aim to broaden the application of DoCL beyond conventional logical solvers, which typically provide limited outcomes such as SAT or UNSAT. By expanding its usage to encompass a wider array of logic code scenarios, we can ensure that DoCL remains effective across diverse problem-solving domains. Besides, through integration with research endeavors focusing on the conversion of natural language into logical code, DoCL stands poised to establish a groundbreaking paradigm in logic-based problem-solving methodologies. Furthermore, by combining LLMs with knowledge retrieval and storage techniques, we look forward to the real-life application of LLM-based logic solvers, which can provide incredible efficiency in simulating complex logic programs.

\bibliography{icse}

\end{document}